\pdfoutput=1
\documentclass[preprint,12pt,authoryear]{elsarticle}

\usepackage{amssymb}
\usepackage{amsmath}
\usepackage{algorithm}
\usepackage{algpseudocode}
\usepackage{subcaption}
\usepackage{tabularray}
\usepackage{hyperref}

\journal{International Journal of Disaster Risk Reduction}

\begin{document}

\begin{frontmatter}



\title{A System for Critical Facility and Resource Optimization in Disaster Management and Planning} 

\author[novateur]{Emmanuel Tung\corref{CorrespondingAuthor}} 
\ead{etung@novateur.ai}

\author[texasam]{Ali Mostafavi}
\ead{amostafavi@civil.tamu.edu}

\author[novateur]{Maoxu Li}
\ead{mli@novateur.ai}

\author[novateur]{Sophie Li}
\ead{sli@novateur.ai}

\author[novateur]{Zeeshan Rasheed}
\ead{zrasheed@novateur.ai}

\author[novateur]{Khurram Shafique}
\ead{kshafique@novateur.ai}

\affiliation[novateur]{organization={Novateur Research Solutions},
            addressline={20110 Ashbrook Pl, Suite 170}, 
            city={Ashburn},
            postcode={20147}, 
            state={VA},
            country={USA}}

\affiliation[texasam]{organization={UrbanResilience.AI Lab, Zachry Department of Civil and Environmental Engineering, Texas A\&M University},
            addressline={3136 TAMU}, 
            city={College Station},
            postcode={77843}, 
            state={TX},
            country={USA}}

\cortext[CorrespondingAuthor]{Corresponding author}

\begin{abstract}
Disruptions to medical infrastructure during disasters pose significant risks to critically ill patients with advanced chronic kidney disease or end-stage renal disease. To enhance patient access to dialysis treatment under such conditions, it is crucial to assess the vulnerabilities of critical care facilities to hazardous events. This study proposes optimization models for patient reallocation and the strategic placement of temporary medical facilities to bolster the resilience of the critical care system, with a focus on equitable outcomes. Utilizing human mobility data from Texas, we evaluate patient access to critical care and dialysis centers under simulated hazard scenarios. The proposed bio-inspired optimization model, based on the Ant Colony optimization method, efficiently reallocates patients to mitigate disrupted access to dialysis facilities. The model outputs offer valuable insights into patient and hospital preparedness for disasters. 
Overall, the study presents a data-driven, analytics-based decision support tool designed to proactively mitigate potential disruptions in access to critical care facilities during disasters, tailored to the needs of health officials, emergency managers, and hospital system administrators in both the private and public sectors.
\end{abstract}

\begin{graphicalabstract}
\includegraphics[width=\textwidth]{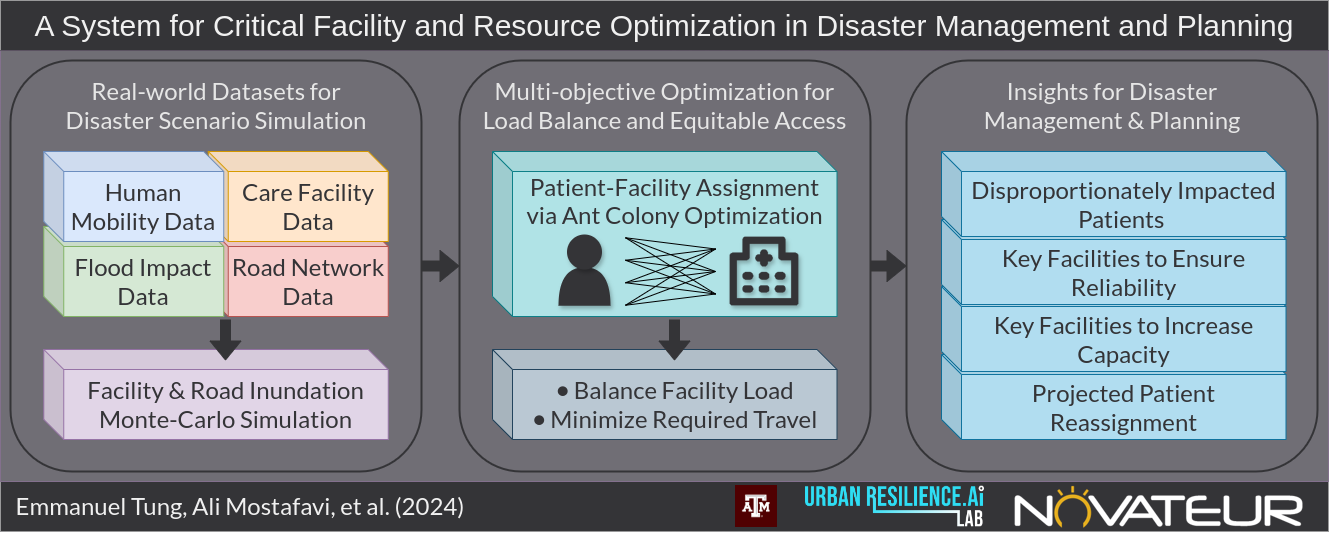}
\end{graphicalabstract}

\begin{highlights}
\item Bio-inspired optimization gives insights for healthcare network disaster resilience
\item Ant Colony Optimization is effective for ensuring equitable access to care facilities
\item Real human mobility data and simulated hazard impacts act as an empirical testbed
\end{highlights}

\begin{keyword}
Critical Facilities \sep Bio-inspired \sep Optimization \sep Disasters \sep Resilience \sep Healthcare Network


\end{keyword}

\end{frontmatter}



\section{Introduction}

Healthcare systems are under immense pressure during disasters, both natural and man-made, leading to a surge in demand for medical services and exacerbating the shortage of healthcare resources in affected areas. This study seeks to develop an equitable optimization framework aimed at enhancing the resilience of critical care networks, with a specific focus on dialysis centers. Resilience, in this context, refers to the healthcare system's capacity to minimize the impacts of disasters while continuing to meet the needs of the population. Dialysis centers are crucial for safeguarding the health of highly vulnerable patients, particularly those with advanced chronic kidney disease or end-stage renal disease. Disruptions in these services during disasters can result in severe, life-threatening kidney failure. 

Patient risk is especially heightened during severe weather events such as hurricanes, floods, or extreme cold, which can lead to the forced closure of these dialysis centers. \citet{lempert13} describes such scenarios as “kidney failure disasters,” where a large number of patients, either on maintenance dialysis or newly diagnosed with acute kidney injury (AKI), face significant health risks due to the unavailability of dialysis services. Historical events like Hurricane Katrina in 2005, Hurricane Gustav in 2008, and Hurricane Sandy in 2012 exemplify the serious challenges faced during such crises. Dialysis services were often preemptively closed in anticipation of storms or were rendered inoperable due to flooding, power outages, and structural damage. These closures forced hospitals to admit evacuated patients, many of whom presented with hyperkalemia in emergency rooms. Despite rapid responses from the renal community, some patients experienced severe health risks and complications due to missed dialysis sessions. Additionally, the uncertainty and disruptions from such events likely had acute and long-term mental health impacts on maintenance dialysis patients.

Despite the critical nature of this issue, the study of disaster-induced disruptions to dialysis centers remains under-explored in healthcare services research. One of the few investigations in this area, conducted by \citet{kaiser21}, assessed the impact of flooding on dialysis centers in Harris County, Texas, during Hurricane Harvey. This study utilized flood maps and FEMA flood zone classifications to evaluate the promximity of dialysis centers to flood-prone areas. However, focusing solely on flood exposure fails to capture the full spectrum of threats to patient access resulting from compromised transportation routes, facility closures, and community disruptions. 

Access to critical care facilities is crucial during disasters \citep{esmalian22, dong20}. Dialysis centers, in particular, are essential to vulnerable populations with chronic illness \citep{yuan23}. Extreme weather events can disrupt access to these critical care facilities due to flooded roads, facility closures, or power and water outages. Patients needing hemodialysis are highly dependent on timely treatment, and even brief disruptions in access can lead to catastrophic outcomes. Extreme weather events, such as Hurricane Katrina in 2005 \citep{bonomini11, vanholder09, kopp07}, Hurricane Gustav in 2008 \citep{kleinpeter08}, Hurricane Harvey in 2017, and Winterstorm Uri in 2021, have led to significant kidney failure disasters due to disrupted access to dialysis centers. These events have underscored the need for a systemic approach to managing patient redistribution across a network of dialysis facilities to ensure all patients receive the care they need, no dialysis center becomes overwhelmed, and travel time and costs are minimized.

\begin{figure}[!ht]
  \includegraphics[width=\textwidth]{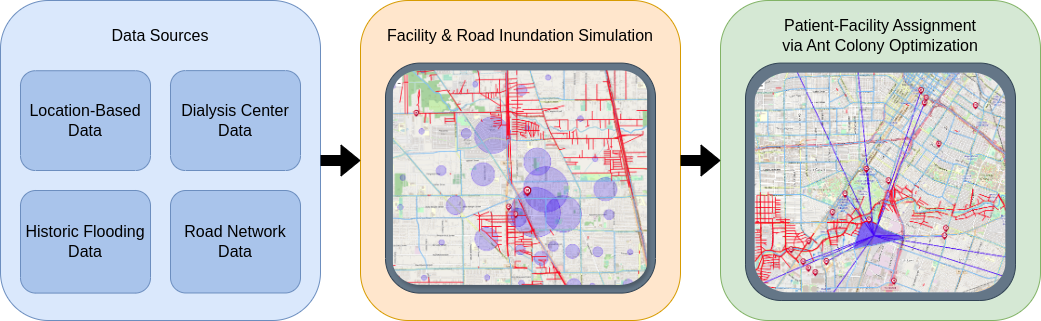}
  \caption{Proposed data-driven methodology for patient-dialysis center assignment during disaster. We source multiple layers of real-world data to inform probabilistic facility \& road closure simulation.}
  \label{fig:overview}
\end{figure}

To address this critical issue, this study presents an optimization methodology based on location data to model patterns of patient visits to dialysis centers during normal times, and to simulate probabilistic disruptions to road networks and facilities using historic flooding data (\autoref{fig:overview}). The model, employing principles from Ant Colony Optimization, optimizes for the competing objectives of facility load balance and travel distance. The results provide insights into facility capacity utilization, identify traffic analysis zones (TAZs) where patients redistribution is necessary, and estimate the average travel distances required of patients.  

The paper is organized as follows: Section 1.1 reviews related work in the field, providing a context for the study. Section 2 details the data preparation and modeling processes employed. In Section 3, we discuss the optimization algorithms and underlying theory, along with the analysis approach. Finally, Section 4 describes the results obtained for the Houston Metro area, using data from Hurricane Harvey.

\subsection{Related Works}
\subsubsection{Access to Critical Services in Disasters}
The current literature on access to critical facilities such as grocery stores, hospitals, and similar services during disasters predominantly employs analytical and simulation-based approaches. These approaches often rely on gravity-type and utility-based measures of access \citep{dong06, islam11, liu14, an15, simini21}. For example, \citet{logan20} proposed a resilience assessment framework that evaluates equitable access to essential facilities like grocery stores and gas stations, measuring changes in the distance for different subpopulations. Similarly, \citet{faturechi14} developed a mathematical model to assess the travel time resilience for transportation networks during disaster, while \citet{dong20} integrated a network percolation measure based on road network topology to evaluate access to healthcare services. However, due to limited data and empirical methods, static distance measures and simulation-based approaches have traditionally been the primary means of evaluating access to critical facilities.

Although the significance of access to critical care facilities, such as dialysis centers, is acknowledged in the literature, most studies focus on service access during normal conditions. For instance, \citet{roserobixby04} employed GIS tools to define the demand-supply system between census population and health facilities, developing an accessibility index based on proximity to the nearest healthcare facility to evaluate spatial access to health care in Costa Rica. Similarly, \citet{jin15} used census data to examine the spatial inequity of access to healthcare facilities in China, creating an index based on travel time from residential buildings to the nearest healthcare facility using least cost path analysis. \citet{mayaud19} extended this approach by evaluating access equity to healthcare facilities in Cascadia, optimizing travel time from census block groups (CBGs) to healthcare facilities using OpenTripPlanner, an open source routing engine. However, these studies largely focus on static spatial configuration of healthcare centers and often overlook the dynamic aspects of access, such as changes in the facility usage patterns due to system disturbances. 

Research on disaster-induced disruptions to critical dialysis centers is particularly sparse. One notable study by \citet{kaiser21} examined the impact of flooding on dialysis centers in Harris County, Texas, during Hurricane Harvey, using flood maps from the incident and FEMA flood zone classifications to evaluate proximity to flood-prone areas. However, this focus on flood exposure alone does not fully capture the array of potential threats to patient access. Flooding can disrupt access through road inundation \citep{redlener12}, facility closures \citep{kaiser21}, and broader community impacts where dialysis-dependent patients reside \citep{du12}.

To mitigate disrupted access to critical care facilities like dialysis centers, two primary strategies have emerged: reallocating patients across the network of regional facilities and establishing temporary facilities to meet demand \citep{muramaki15}. Several optimization methods have been proposed to address patient and medical resource allocation challenges \citep{fiedrich00, minciardi09, revelle95, sun14, tsai22, ye22, yi07, ma13}. For instance, \citet{tsai22} applied linear programming models during a dengue fever epidemic to minimize total travel distances for patient allocation. \citet{ma13} developed an integer linear programming (ILP) model to optimize bed allocation and financial outcomes for hospitals, considering constraints such as bed capacity and occupancy. Similarly, \citet{sun14} addressed patient and resource allocation during a pandemic influenza outbreak, focusing on minimizing both total and maximum travel distances. \citet{ye22} proposed a multi-objective planning method for patient allocation during major epidemics, taking patient severity into account. \citet{mosallanezhad21} introduced a multi-objective model for personal protective equipment allocation during the COVID-19 pandemic, aiming to satisfy demand while minimizing total costs.

In the context of disaster response, \citet{minciardi09} developed a mathematical model to support optimal resource allocation before and during natural hazard emergencies. \citet{revelle95} addressed emergency room location optimization to ensure maximum demand coverage. \citet{fiedrich00} explored resource allocation strategies to minimize casualties during the initial search and rescue phase following major earthquakes. \citet{yi07} created an integrated location-distribution model for selecting temporary emergency centers to optimize post-disaster medical coverage and resource distribution. \citet{gulzari21} focused on the strategic location of temporary health facilities and the allocation of health professionals, incorporating telemedicine into earthquake response to minimize unmet healthcare demand.

Despite significant advancements, there remains limited research on the impacts of disasters on access to critical dialysis centers. The few studies that do exist, such as \citet{kaiser21}, primarily examine flood exposure without fully considering the broader vulnerabilities patients face during disasters. This gap highlights the need for more comprehensive approaches that account for dynamic disruptions to access and infrastructure.

Recent advancements in data acquisition and data analytics have opened new avenues for examining human mobility during community-level disruptions and access to essential facilities during disasters \citep{esmalian22, yuan23}. Although mathematical models have been applied to patient and resource allocation problems, they often overlook the complexities introduced by potential infrastructure disruptions following hazard events. Many studies assume that existing facilities remain operational post-disaster \citep{mete10, mohammadi16, rabbani16}, an assumption that is often unrealistic, given the potential for severe infrastructure damage \citep{galindo13}.

\subsubsection{Ant Colony Optimization}

Ant System, introduced by \citet{colorni91}, was the first Ant Colony Optimization (ACO) algorithm, originally designed to solve the Traveling Salesman Problem. Since then, numerous variants and extensions, such as the Elitist Ant System, Max-Min Ant System, and Ant Colony System, have been developed to enhance performance \citep{stutzle04}. The Ant Colony System (ACS) \citep{dorigo97} has been adapted for both single-objective resource assignment \citep{rezende18} and multi-objective resource allocation problems \citep{pham17}.

Although the ACS algorithm theoretically converges to an optimal solution, as noted by \cite{stutzle04}, this convergence is not always practical in real-world applications. ACS is capable of solving both static problems (with unchanging characteristics) as well as dynamic problems (where characteristics may change during runtime). One of the strengths of ACS lies in its distributed nature, which allows for parallel implementations. Multiple surveys and comparative studies \citep{afzal19, kalra15} have explored the advantages and disadvantages of various meta-heuristic algorithms, including ACO, in the context of resource allocation in cloud computing. However, further research is needed to determine the optimal application of these algorithms for achieving the best results in both solution quality and runtime.

In resource allocation, the classical linear Transportation Problem (TP) - seeking to assign resources while minimizing transportation costs - can be optimally solved in polynomial time \citep{schrenk11}. The TP formulation only addresses a single objective, transportation cost. However, patient allocation during disaster is a multi-objective optimization problem, which involves both minimizing transportation cost (e.g., travel time) as well as balancing load across many dialysis centers. Meta-heuristic algorithms like Ant Colony Optimization \cite{pham17} and Firefly Optimization \cite{abedi22} have been proposed as methods for balancing load in cloud computing resource allocation by minimizing the standard deviation or variance of resource loads. In this study, we adapt and apply one such meta-heuristic algorithm, the Ant Colony System, to address a bi-objective problem that involves minimizing transportation costs and balancing hospital loads, with a focus on assigning patients to their preferred hospital wherever possible within a two-stage optimization framework.

\section{Material and methods}
\subsection{Overview}

Our methodology overview is as follows. See \autoref{fig:overview} for a graphical overview.


\begin{itemize}
    \item We sourced real-world data about the city of Houston, Texas regarding its dialysis center (hereafter ``hospital'') locations, traffic analysis zone (TAZ) centroid locations, road network, historical geographic flooding patterns, and dialysis patient (hereafter ``patient'') location activity surrounding the time period of Hurricane Harvey (August-September 2017).
    \item We found that the data suggest unorganized and inefficient patient-hospital reassignment during the time of Hurricane Harvey, which inspired this work.
    \item We used the real-world data to inform the conditions of (100) probabilistic simulation scenarios of road and hospital inundation during a hurricane. 
    \item We adapted and applied the Ant Colony System algorithm to find Pareto-optimal solution sets the patient-hospital assignment problem, optimized according to two competing objective functions: minimizing travel distance cost for patients and balancing the patient-induced load across all hospitals. We further processed the algorithm output to create practical insights about the simulation solutions, both individually and in aggregate. 
    \item We conducted analysis on the results of the simulations to demonstrate the concept of organized patient reassignment due to hospital closure and road flooding during a hurricane. We generated insights about recommending key hospitals for an increase in capacity, recommending key hospitals for an increase in reliability, classifying TAZs as being at risk of needing to drive significantly farther in case of a hurricane, etc. We visualized and exhibited these insights by developing a graphical user interface for displaying our findings on an interactive map.
\end{itemize}

\subsection{Data Collection and Processing}

\subsubsection{Study Context}

This study examines the access to dialysis centers by TAZs in Harris County, Texas. Harris County, home to the fourth largest city Houston in the United States, has seen rapid population growth over past decades. Harris County is prone to flood and hurricane due to its coastal location, burgeoning urban development, as well as lack of flood control infrastructure development in parallel with the development and population growth \citep{dong20}. Harris County also possesses a diverse population with varying sociodemographic characteristics, which provides us with a great testbed to demonstrate the application of the proposed optimization model.

\subsubsection{Data Description and Processing}

Four datasets have been used in this study including location-based data, dialysis center facility data, 100-year and 500-year flood plains, and road network data (see \autoref{table:data}).

Each dataset is preprocessed as follows:

\textbf{Location-based Data}: 1) Identify if a user has visited a dialysis center by detecting the overlap of the dialysis center polygon and user stop points; 2) Aggregating number of dialysis center visits of a user for each day; 3) With user ID, determine the different TAZs each dialysis center receives patients from.

\textbf{Dialysis Center Data}: 1) Extract geo-locations of dialysis centers; 2) Identify each dialysis center's capacity using the number of stations.

\textbf{Historic Flooding Data}: 1) Overlap the flood
inundation maps with the road network and dialysis centers to determine the likelihood of inundation during 100-year and 500-year events.

\textbf{Road Network Data}: Extract topology of the road network in the study area and length of road segments.


We have sourced a private mobility dataset which combines different sources like Bluetooth, IoT, and Wi-Fi Signals at the device level to provide accurate geographical coordinates and high-quality mobility data. To understand the patterns of visits from different TAZs to different dialysis centers, we harnessed this device-level mobility data, which includes the information of home census block groups, census block groups of stay, and visits for each anonymized device to different facilities. In this study, we aggregated the visits by devices of each TAZ to each facility during each week to obtain the weekly TAZ-facility visits. The aggregated visits serve as the normal period demand on each facility from each TAZ. This baseline visits to each facility was used to estimate the capacity of each facility and implementing the optimization model as explained in the next section.


\begin{table}
\centering
\begin{tblr}{ |p{2.6in}|p{2.6in}| } \hline
 \SetCell[c=2]{c}{\textbf{Location-based Data}} \\ \hline  \SetCell[r=2]{l}{\textbf{Description:} Aggregated data sources providing the Stop Points by Device. Each row represents a trip from a TAZ to different dialysis centers.} & \textbf{Derived Information:} Daily trips from TAZs to dialysis centers \\ & \textbf{Source:} Private \\ \hline\hline \SetCell[c=2]{c}{\textbf{Dialysis Center Data}} \\ \hline \SetCell[r=2]{l}{\textbf{Description:} Data for 124 hemodialysis clinics in Harris County, such as location and number of stations.} & \textbf{Derived Information:} Locations and dialysis capacity \\ & \textbf{Source:} Medicare.gov \\ \hline\hline \SetCell[c=2]{c}{\textbf{Historic Flooding Data}} \\ \hline \SetCell[r=2]{l}{\textbf{Description:} The flood inundation map of 100-year and 500-year plains by Federal Emergency Management Administration (FEMA).} & \textbf{Derived Information:} Inundation / disruption likelihood of roads and facilities \\ & \textbf{Source:} FEMA (2018) \\ \hline\hline \SetCell[c=2]{c}{\textbf{Road Network Data}} \\ \hline \SetCell[r=2]{l}{\textbf{Description:} Open geographic database including roads, buildings, addresses, shops and businesses, points of interest, railways, trails, transit, land use and natural features, etc.} & \textbf{Derived Information:} Road network topology \& length of road segments \\ & \textbf{Source:} Open Street Map \\ \hline
\end{tblr}
\caption{Data description, source, derived information from each dataset.}
\label{table:data}
\end{table}

\subsubsection{Hazard Scenarios and Simulation of Disruptions} \label{subsection:scenario_gen1}

To simulate disruption scenarios, we used the 100-year and 500-year flood plain data to calculate the likelihood of inundation of roads and dialysis center facilities during a 100-year extreme weather event. If a road segment or a facility is located in a 100-year flood plain, the probability of inundation in a 100-year flood event is 1. If part of a road or a facility is located in a 500-year flood plain, the likelihood of inundation in 100-year event is 0.2. Facilities and road segments located outside the 100-year and 500-year flood plains have likelihood of inundation of zero. Based on these estimated inundation likelihoods, we implemented 100 probabilistic scenarios using Monte-Carlo simulation in which different road segments and facilities are inundated (roads disrupted and facilities go out of service) and implemented the optimization model for each scenario individually and collectively. For each scenario, the model determined the number of patients served by each facility, the TAZs whose patients need to be redistributed, the capacity utilization of each facility, and the distance travelled by patients in each TAZ. 

\subsection{Data Preparation \& Modeling} \label{subsection:scenario_gen2}
We use the data sources above to model (100) discrete, week-based scenarios of resultant road and hospital states due to hurricane flooding. In particular, each scenario is determined by a hospital-TAZ distance cost matrix and a list of hospital closures (a “closure” means the hospital will have zero capacity for the scenario). The other factors (number of patients in each TAZ, preferred hospitals of each patient, capacity of each hospital when operational) remain static across all scenarios. 

Both the hospital-TAZ distance cost matrix and the hospital closure list are derived from assigning floodplain areas a certain probability of flooding based on its historical category (e.g., “100-year floodplain”). Because considering flooded roads as $\textit{unusable}$ made solutions impossible, we consider flooded roads to be assigned (10) times the normal distance cost, in order to account for the additional difficulty of driving over flooded roads. The hospital-TAZ distance cost matrix, then, is the distance of shortest path in the artificial road network for each hospital-TAZ combination. This distance cost matrix is re-calculated for each scenario using a many-to-many shortest-path algorithm, as the shortest-path route and total distance changes from scenario to scenario. 

The static factors are derived from pre-hurricane data, which is assumed to be representative of normal operation. Each recorded visit is assumed to be a distinct patient with the visited hospital being the patient’s preferred hospital. A patient is assumed to require one visit per calendar week, and the number of patients remains constant. The maximum weekly capacity of each hospital is assumed to be 4/3 of its weekly visit count in the pre-hurricane data. Hospitals and TAZs with no visits during this period were excluded from consideration, leaving each scenario with (95) hospitals, (1750) TAZs, and (18002) patients.

Each scenario’s data is ingested by a Python preprocessing stage, modeled and optimized with the Ant Colony System (ACS) algorithm in Netlogo, and then postprocessed in Python. The key component of the preprocessing stage is the pre-assignment phase, in which patients whose preferred hospital is not closed are “pre-assigned” to those hospitals. The justification for this is that patients would always prefer visiting their preferred hospital because of familiarity with the medical staff, insurance constraints, etc. This creates a partial solution and leaves the remaining optimization problem of how to best assign those patients who are unable to visit their preferred hospital due to closure. It is the optimization algorithm’s function to assign these remaining patients based on given multiple objective functions and corresponding heuristics. The algorithm regards each hospital's remaining capacity after pre-assignment as the capacity of that hospital. The postprocessing stage chiefly merges the partial solution (from pre-assignment) and the optimized solution (from the algorithm).  

\section{Theory and calculation}
\subsection{Optimization Problem Formulation} \label{subsection:formulation}
For a given scenario $\mathbf{S} = \{\mathbf{c}, \mathbf{o}, \mathbf{D}\}$, derived by \ref{subsection:scenario_gen1} and \ref{subsection:scenario_gen2}, we define the optimization problem as follows.
\begin{itemize}
  \item Let $\mathbf{c} \in (\mathbb{Z}_{>0})^{n_h}$ describe the weekly patient capacity of each hospital $i$ ($n_h$ hospitals in total). 
  \item Let $\mathbf{o} \in (\mathbb{Z}_{>0})^{n_t}$ describe the weekly patient populations of each TAZ $j$ ($n_t$ TAZs in total).
  \item Let $\mathbf{D} \in (\mathbb{R}_{>0})^{n_h \times n_t}$ describe the matrix of minimum distance costs between each hospital $i$ and TAZ $j$.
  \item It is assumed that $\sum_{i} c_i \geq \sum_{j} o_j$, such that there exist solutions for $\mathbf{S}$
\end{itemize}
Then, we attempt to find a set $\mathbf{\Gamma}$ of Pareto-optimal solutions according to the following objectives and constraints. Let a single solution be denoted as $\mathbf{A}$.
\begin{itemize}

\item Let $\mathbf{A} \in (\mathbb{Z}_{\geq0})^{n_h \times n_t}$ be a matrix of patient assignments between each hospital $i$ and TAZ $j$.

\item Minimize objective function $f_0^{\mathbf{S}}$ - the total distance cost all patients must travel:
\begin{equation} 
\text{Minimize: } f_0^{\mathbf{S}}(\mathbf{A}) = \langle \mathbf{A}, \mathbf{D} \rangle_F \label{eq:obj0}
\end{equation}

\item Minimize objective function $f_1^{\mathbf{S}}$ - the corrected sample standard deviation of the relative unused capacities of hospitals:
\begin{equation}
\begin{aligned}
\text{Minimize: } f_1^{\mathbf{S}}(\mathbf{A}) & = \sqrt{\frac{1}{n_h - 1} \sum_{i = 1}^{n_h} (r_i - \bar{\mathbf{r}}) ^ 2}\\ \label{eq:obj1}
\text{where } \mathbf{r} & = \text{diag} ^ {-1}(\mathbf{c})(\mathbf{c} - \mathbf{u})\\
\text{and } u_i & = \sum_{j} A_{i, j} \quad \forall i \in \{1, \dots, n_h\}
\end{aligned}
\end{equation}

\item Subject to - no hospital may exceed capacity:
\begin{equation*}
c_i - \sum_{j} A_{i, j} \geq 0 \quad \forall i \in \{ 1, \dots, n_h \} 
\end{equation*}

\item Subject to - no TAZ may have any patients without treatment:
\begin{equation}
o_j - \sum_{i} A_{i, j} = 0 \quad \forall j \in \{ 1, \dots, n_t \}
\label{eq:c1}
\end{equation}
\end{itemize}
In this way, the optimization problem is modeled as a many-to-one patient-to-hospital assignment problem. The optimization algorithm generates many valid solutions, of which the Pareto-optimal solutions (according to $f_0^{\mathbf{S}}$ and $f_1^{\mathbf{S}}$) are kept.

\subsection{Optimization Algorithm}
The first Ant Colony Optimization algorithm, Ant System \cite{colorni91}, was first applied to the Traveling Salesman Problem. We adapt the Ant Colony System (ACS) \cite{dorigo97} algorithm to this multi-objective optimization problem. We do not explicitly model this optimization problem in the form of a Traveling Salesman Problem instance, but rather adopt the driving principles of ACS, those being the usage of heuristics and pheromones to guide the exploration of the search space, to fit to the needs of the problem. 

\subsubsection{State Transition Rule}
In each iteration, each artificial ant incrementally builds a valid solution by repeatedly selecting a random unassigned patient and assigning a hospital based on the state transition rule. In other words, an ant assigns individual patients to hospitals in a random order until all patients are assigned.
\\
\\
Let $\mathbf{A}$ be a partially completed solution within iteration $t$ for scenario $\mathbf{S}$, meaning it is a valid solution with the exception that \autoref{eq:c1} may not hold. The non-deterministic state transition rule for assigning patient $k$ to hospital $i$, then, is as follows:
\begin{equation}
i = 
\begin{cases}
  \arg \max_{i' \in \mathbf{\mu}}\{[p_t(i', k)][h^{\mathbf{S}}_{\mathbf{A}}(i', k)]^\beta\}, & \text{if $q < q_0$}.\\
  \text{Sample by \autoref{eq:str1} probability distribution}, & \text{otherwise}.
\end{cases}
\label{eq:str}
\end{equation}

\begin{equation}
P_{i, k} = 
\begin{cases}
  \frac{[p_t(i, k)][h^{\mathbf{S}}_{\mathbf{A}}(i, k)]^\beta}{\sum_{i' \in \mu} [p_t(i', k)][h^{\mathbf{S}}_{\mathbf{A}}(i', k)]^\beta}, & \text{if $i \in \mu$}.\\
  0, & \text{otherwise}.
\end{cases}
\label{eq:str1}
\end{equation}

\begin{equation}
\mu = 
\left\{
i \in \{ 1, \dots, n_h \} |
c_i - \sum_{j} A_{i, j} > 0
\right\}
\label{eq:str2}
\end{equation}
\\
where:
\begin{itemize}
    \item $p_t(i', k)$ is the pheromone value for hospital $i'$ and patient $k$ on iteration $t$ (\ref{subsection:ph})
    \item $h_{\mathbf{A}}^{\mathbf{S}}(i', k)$ is the heuristic evaluation for hospital $i'$ and patient $k$ (\ref{subsection:h})
    \item $\beta$ is the heuristic importance parameter
    \item $q$ is a uniformly random sampled value from the range $[0, 1)$, sampled each time the rule is applied
    \item $q_0$ is the exploitation rate
    \item $P_{i, k}$ is the probability of assigning patient $k$ to hospital $i$ given that $q \geq q_0$
    \item $\mu$ is the set of hospitals with positive unused capacity given partial solution $\mathbf{A}$
\end{itemize}
In words, the state transition rule may exploit its current knowledge by assigning the patient to the most attractive hospital according to the joint pheromone and heuristic evaluation (\autoref{eq:str}). Or the rule may explore by randomly sampling a hospital according to the distribution which is weighted by the joint pheromone and heuristic evaluation (\autoref{eq:str1}). An ant only considers legal hospitals for the patients, meaning hospitals with zero remaining capacity are excluded (\autoref{eq:str2}).
\\
\\
At the time that patient $k$ is assigned to hospital $i$ within partial solution $\mathbf{A}$, $\mathbf{A}$ is updated by way of incrementing $A_{i, j}$ for patient $k$ originating from TAZ $j$.

\subsubsection{Pheromone Update Rule} \label{subsection:ph}
Arguably the core principle of the Ant Colony System algorithm is the shared use of pheromone values to increase or decrease attractiveness of a solution component based on the solutions constructed by other ants. Each solution component, which is a single patient-hospital assignment, is initially given the same value, denoted by:
\begin{equation}
p_0(i, k) = (n_a l)^{-1}, \quad
\forall i \in \{ 1, \dots, n_h \}, \,
\forall k \in \{ 1, \dots, \sum_{j} o_j \}
\label{eq:ph0}
\end{equation}
where $n_a$ is the number of ants, and $l$ is the single objective function cost $f_0^{\mathbf{S}}(\mathbf{A})$ for some trivial, poor-quality solution $\mathbf{A}$. A similar formula for the initial pheromone has historically been found to be more effective than setting the initial pheromone to zero, although any very rough approximation for this formula is sufficient \citep{dorigo97, pham17, rezende18}. Therefore, we use the following upper-bound approximation: $l = n_a \; \text{max}\{\mathbf{D}\}$, so $p_0(i, k) = (n_a^2 \, \text{max}\{\mathbf{D}\})^{-1}$. We hereafter denote the initial pheromone as $p_0$.
\\
\\
Beyond the initial pheromone value $p_0(i, k) = p_0$, the pheromone mapping is updated locally (once each time a solution component is chosen) and globally (once each iteration). In the ACS algorithm, the local update is used to decrease pheromones for solution components which have been used in the constructed solutions of other ants, whereas the global update is used to increase pheromones for solution components present within the Pareto-optimal solution set. Therefore, the local update encourages exploration, and the global update encourages exploitation, and their associated pheromone decay rates must be balanced accordingly.
\\
\\
The local pheromone update rule is applied each time a patient $k$ is assigned to hospital $i$:
\begin{equation} \label{eq:ph1}
\begin{aligned}
p_{t}(i, k) & \leftarrow (1 - \alpha_l) \, p_t(i, k) + \alpha_l \, \Delta p_t(i, k) \\
\Delta p_t(i, k) & = p_0
\end{aligned}
\end{equation}
where $\alpha_l$ is the local pheromone decay rate. 
\\
\\
The global pheromone update rule is applied at the end of each iteration for each patient $k$ originating from TAZ $j$ assigned to hospital $i$ in within each solution in the running Pareto-optimal solution set $\Gamma'$:
\begin{equation}
\begin{aligned}
\Gamma_{i, k}' & = \{\mathbf{A}' | \mathbf{A}' \in \Gamma', \; A_{i, j}' > 0\} \\
p_t(i, k) & \leftarrow (1 - \alpha_g) \, p_t(i, k) + \alpha_g \, \Delta p_t(i, k); \quad \forall \Delta p_t(i, k) \in \{ (f_{0}^{\mathbf{S}}(\mathbf{A}'))^{-1} | \mathbf{A}' \in \Gamma_{i, k}' \} \\
p_{t + 1}(i, k) & \leftarrow p_t(i, k)
\end{aligned}
\label{eq:ph2}
\end{equation}
where $\alpha_g$ is the global pheromone decay rate.

\subsubsection{Heuristic Functions} \label{subsection:h}
The heuristic function $h^{\mathbf{S}}_{\mathbf{A}}(i, j)$ determines the heuristically evaluated attractiveness score of assigning a patient from TAZ $j$ to hospital $i$, given partial running solution $\mathbf{A}$ and scenario $\mathbf{S}$. It is a deterministic function which is used to guide the exploration of the search space. As there are multiple competing objective functions, there are multiple corresponding heuristic functions, one for each objective function. Each heuristic $h$ function has a respective $\lambda$ hyperparameter which affects resultant distribution of the heuristic function.
\\
\\
Heuristic function $h_0$, corresponding to objective function $f_0$, aims to assign higher values to lower distance costs, and vise versa. It is a monotonically decreasing function, given by:
\begin{equation*} 
h_0 : [0, 1] \rightarrow [m, 1], \, m > 0; \quad h_0(d) = m ^ {d ^ {\gamma_0}} 
\end{equation*}
Where $h_0$ is the heuristic function value, $d$ is the normalized distance cost, $m = 0.02$ is the minimum heuristic value, and $\gamma_0$ is a constant which defines the shape of the heuristic curve. Lowering $\gamma_0$ tends to increase favorability of distance costs very close to zero, whereas raising $\gamma_0$ tends to flatten out the extremes with a sharper drop-off in the mid-range.
\\
\\
Heuristic function $h_1$, corresponding to objective function $f_1$, aims to assign higher values to higher relative remaining capacities. It is a monotonically increasing function, given by:
\begin{equation*}
h_1 : [0, 1] \rightarrow [0, 1]; \quad h_1(r) = r ^ {\gamma_1}
\end{equation*}
Where $h_1$ is the heuristic value, $r$ is the relative unused capacity, and $\gamma_1$ is some constant which defines the shape of the heuristic curve. Lowering the $\gamma_0$ constant tends to spread favorability across the domain more equally, whereas raising the $\gamma_0$ constant tends to concentrate favorability towards higher relative unused capacity.
\\
\\
It should be noted that $h_0$ is static whereas $h_1$ is dynamic. Therefore, $h_0$ is computed only once for each TAZ-hospital pair, and $h_1$ is re-calculated as necessary.
\\
\\
Let $\mathbf{A}$ a partially completed solution, meaning it is a valid solution with the exception that \autoref{eq:c1} may not hold. The overall heuristic evaluation $h^{\mathbf{S}}_{\mathbf{A}}(i, k)$ for a patient $k$ from TAZ $j$ being assigned to hospital $i$ for a partially completed solution $\mathbf{A}$ is given by the product of $h_0$ and $h_1$:
\begin{equation*}
h^{\mathbf{S}}_{\mathbf{A}}(i, k) = h_0\left(\frac{D_{i, j}}{\text{max}\{\mathbf{D}\}}\right) \; h_1\left(\frac{c_i - \sum_{j'} A_{i, j'}}{c_i}\right) 
\end{equation*}

\subsubsection{Algorithm Configuration}
The ACS algorithm has many potential choices for each parameter value, objective function, and heuristic function. The objective functions determine which solutions are to be kept as part of the Pareto-optimal solution set, which is the output of the algorithm. The heuristic functions help guide the exploration of the search space by specifying a mapping from a measurement to an attractiveness score. The parameters (e.g., local pheromone decay, exploitation rate, etc.) also help to guide the exploration of the search space and are dependent on the subject data.

We use Netlogo’s parallel simulation functionality to optimize each scenario with (9) different runs. Each run has parameters in common (\ref{app1}), which were found to be effective and efficient (increasing iteration count and ant count, for example, has not been found to noticeably increase the best solutions’ quality).

Then, each run differs only in the values for $\gamma_0$ and $\gamma_1$ for heuristic functions $h_0$ and $h_1$. The (9) combinations are constructed
as: $(\gamma_0, \gamma_1) \in \{\frac{1}{2}, 1.0, 2.0\} \times \{\frac{1}{3}, 1.0, 3.0\}$.
The shapes of the heuristic curves implicitly define the goals of each heuristic and the prioritization of each heuristic in relation to the other, effectively acting as an embedded weighting scheme for the opposing objective functions. 

Finally, for each scenario, we obtain the universally Pareto-optimal solution set by properly merging the globally Pareto-optimal solution sets from each run. 

\subsubsection{Algorithm Operation and Analysis}
The following pseudocode (Algorithm \ref{pseudocode}) describes the high-level operation of the ACS algorithm as it is applied to the patient-hospital assignment problem. Recall that the scenarios are derived by \ref{subsection:scenario_gen1} and \ref{subsection:scenario_gen2}.

\begin{algorithm}
\caption{ACS for patient-hospital assignment} \label{pseudocode}

\hspace*{\algorithmicindent} \textbf{Hyperparameters: } $n_i, n_a, q_0, \alpha_l, \alpha_g, \beta, \gamma_0, \gamma_1$ \\
\hspace*{\algorithmicindent} \textbf{Input: } Scenario $\mathbf{S} = \{\mathbf{c}, \mathbf{o}, \mathbf{D}\}$, $\mathbf{c} \in (\mathbb{Z}_{>0})^{n_h}$, $\mathbf{p} \in (\mathbb{Z}_{>0})^{n_t}$, $\mathbf{D} \in (\mathbb{R}_{>0})^{n_h \times n_t}$ \\
\hspace*{\algorithmicindent} \textbf{Output: } Pareto Set $\mathbf{\Gamma}$

\begin{algorithmic}[1]

\State $\mathbf{\Gamma} \gets \{  \}$ \hfill (Initialize Pareto set)
\State $n_o \gets \sum_{j} o_j$ \hfill (Calculate number of patients)
\State Calculate initial pheromone by \autoref{eq:ph0}
\State
\For{$t \gets 1$ to $n_i$} \hfill (For each iteration)
    \State
    \For{$a \gets 1$ to $n_a$} \hfill (For each ant, i.e., each new solution)
        \State $\mathbf{A} \gets \mathbf{0}^{n_h \times n_t}$ \hfill (Initialize a new solution)
        \State
        \For{$k \gets $ random permutation of $1$ to $n_o$} \hfill (For each patient)
            \State Assign hospital $i$ for patient $k$ by \autoref{eq:str}
            \State Increment $A_{i, j}$ for TAZ $j$ containing patient $k$
            \State Local pheromone update by \autoref{eq:ph1}
        \EndFor
        \State
        \State Calculate objective function costs for $\mathbf{A}$ by \autoref{eq:obj0}, \autoref{eq:obj1}
        \State Update Pareto set $\mathbf{\Gamma}$ given new solution $\mathbf{A}$
    \EndFor
    \State
    \State Global pheromone update by \autoref{eq:ph2}
\EndFor
\State
\State \Return $\mathbf{\Gamma}$ \hfill (Return Pareto set)
\end{algorithmic}
\end{algorithm}

With caching and updating, the algorithm runtime is given by $O(n_i^2 \cdot n_a^2 \cdot n_h \cdot n_o)$ where $n_o = \sum_j o_j$, the total number of patients. In the worst case, the size of the Pareto set $\mathbf{\Gamma}$ is bounded above by $n_i \cdot n_a$. However, if we assume that $|\mathbf{\Gamma}|$ is bounded above by a constant (as is generally the case in practice), then the runtime is instead given by $O(n_i \cdot n_a \cdot n_h \cdot n_o)$.
Note that the runtime scales linearly with the number of hospitals or the number of patients. The runtime can be drastically reduced in practice by using a pre-computed $\textit{candidate list}$, as suggested in \citet{dorigo97}, which contains the $n_c$ closest hospitals to each patient; an ant then only considers assigning patients to a hospital from the candidate list, unless there are no available options left within it. This type of modification may increase speed at the cost solution quality.

\subsection{Postprocessing \& Scenario Aggregation Procedure} \label{subsection:postprocess}
From our various data, we have constructed (100) scenarios which each determine a distinct configuration of the optimization problem. For each configuration, our process conducts (9) different runs of applying the ACS algorithm, merging the globally Pareto-optimal solutions from each run into a universally Pareto-optimal solution set for that scenario. 

Each solution for each scenario is a discrete assignment of patients from TAZs to hospitals. We postprocess the algorithm output to give additional data about each solution. Namely, from each solution and scenario's data, we derive the following data:
\begin{itemize}
    \item The travel distance cost each patient would incur for a visit to their assigned hospital.
    \item The absolute and relative unused capacity of each hospital.
    \item For closed hospitals, which hospitals their patients are reassigned to, and in what quantities (hereafter the ``hospital-hospital patient re-assignment matrix'').
    \item Each hospital's status as determined by its relative unused capacity; for those that are not closed, a hospital is ``underused'' if unused capacity is greater than $50\%$, ``stressed'' if unused capacity is less than $10\%$, and ``ideal'' otherwise.
    \item Each TAZ's status as determined by the average travel distance cost for its patients according to the solution and scenario; if this average is higher than the average travel distance cost for its patients according to the pre-hurricane data, then the TAZ is considered to have the ``mobility risk'' status for its patients.
\end{itemize}

For each scenario, we then aggregate the universal Pareto-optimal solution set into one aggregate result by averaging the solution matrices. This aggregate solution may be interpreted as the expected number of patients that each TAZ would send to each hospital. Therefore, this aggregate solution may include rational numbers, not necessarily assigning whole numbers of patients. The additional data derived from each solution is also taken into a scenario-wide aggregate by averaging or recording the probability, as appropriate for each metric. Lastly, a similar aggregation procedure may be applied to each scenario-wide aggregate solution for an all-scenario aggregate. 

\section{Results}

\subsection{Pre-hurricane and During-hurricane Baseline}

We use the data from August 10th-16th as the pre-hurricane data representing patient-related and hospital-related figures for Houston, Texas before the 2017 Hurricane Harvey. We use August 25-31st for the period of time during Hurricane Harvey.

\begin{figure}[!ht]
    \centering
    \begin{subfigure}{0.49\textwidth}
        \centering
        \includegraphics[width=0.99\linewidth]{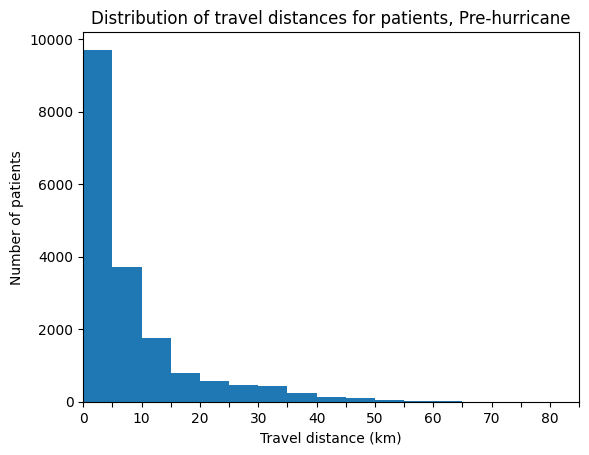}
        \caption{Patient travel distances, Pre-hurricane}
        \label{fig:pre_hurricane_distance}
    \end{subfigure}
    \begin{subfigure}{0.49\textwidth}
        \centering
        \includegraphics[width=0.99\linewidth]{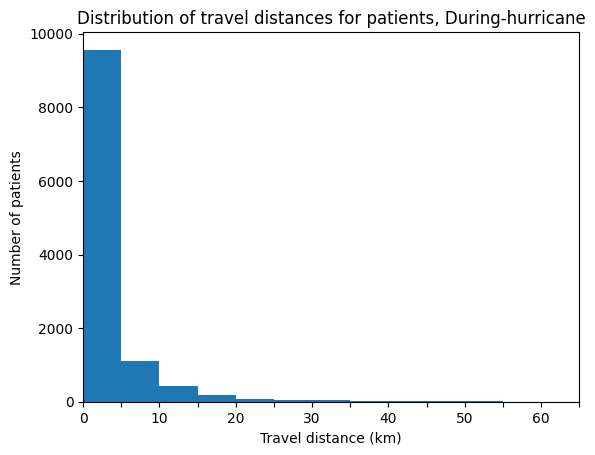}
        \caption{Patient travel distances, During-hurricane}
        \label{fig:during_hurricane_distance}
    \end{subfigure}
    \vskip\baselineskip
    \begin{subfigure}{0.49\textwidth}
        \centering
        \includegraphics[width=0.99\linewidth]{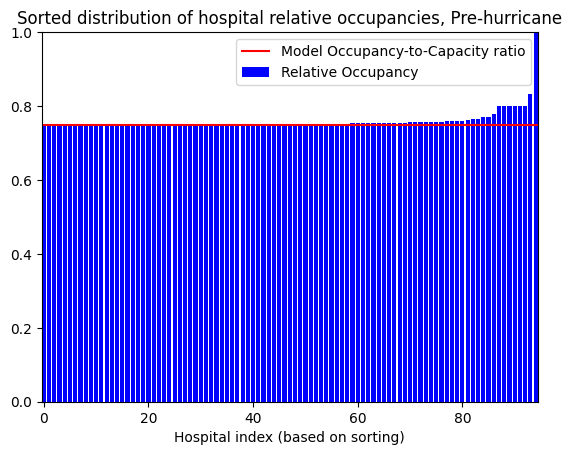}
        \caption{Hospital relative occupancies, Pre-hurricane}
        \label{fig:pre_hurricane_relative_occupancy}
    \end{subfigure}
    \begin{subfigure}{0.49\textwidth}
        \centering
        \includegraphics[width=0.99\linewidth]{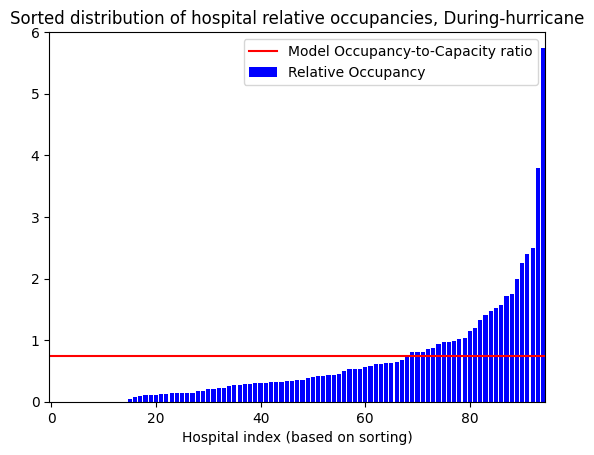}
        \caption{Hospital relative occupancies, During-hurricane}
        \label{fig:during_hurricane_relative_occupancy}
    \end{subfigure}
    \caption{Pre-hurricane (first column) \& During-hurricane (second column) statistics}
\end{figure}

In \autoref{fig:pre_hurricane_distance}, we see the ground truth distribution of distances patients are traveling to visit their preferred hospital before the hurricane, according to our patient mobility data. The maximum distance traveled by any patient is $\approx83.9$ kilometers, which is not far from the maximum TAZ-to-hospital travel distance of $\approx99.2$ kilometers. Out of (18002) patients, slightly more than half are driving less than (5) kilometers to their preferred hospital. 

This is in contrast to \autoref{fig:during_hurricane_distance}, which shows the same data from around the time of the hurricane. It is seen that there are far fewer visits with long ($>5$km) travel distances.

In \autoref{fig:pre_hurricane_relative_occupancy}, we see the distribution of the relative occupancy (number of visits divided by capacity) of hospitals before the hurricane. Our model estimates hospital capacities by taking the pre-hurricane visit counts of each hospital and multiplying by a factor of $4 / 3$; in other words, we assume that all hospitals are operating at a normal occupancy which is proportional to their maximum capacity by a constant factor. Therefore, we might expect to see all hospitals in this plot attain the model occupancy-to-capacity ratio of $3 / 4 = 0.75$. However, because we round down after multiplying by the $4 / 3$ and a few hospital visit counts are quite small (i.e., single-digit), this introduces the aspect that not all hospitals can achieve $0.75$ relative occupancy before the hurricane, although the vast majority do. 

This is in constrast to \autoref{fig:during_hurricane_relative_occupancy}, which again shows the same data from around the time of the hurricane. It is seen that for the same (95) hospitals as in the pre-hurricane data, these hospitals are being severely overused (i.e., occupancy is greater than modeled capacity), underused, or closed (unused).

It is noted that modeled during-hurricane hospital visit counts are more erratic day-to-day than pre-hurricane figures, with greater peaks and lower minimums. The hospital ``Davita - North Houston Dialysis'' has the greatest relative occupancy, with an estimated usage $5.75$ times greater than its modeled capacity. The largest hospital which exceeded capacity had a capacity of $1229$ weekly visits; this hospital has a an occupancy-to-capacity ratio of $\approx 1.52$ and supported $637$ patients over capacity. 

In summary, our hurricane data show that fewer patients are visiting dialysis centers during the hurricane, and those patients that are visiting dialysis centers are more concentrated on minimal travel distance categories; and, many patients are no longer visiting their preferred hospital and hospitals are being severely underused or overused. Therefore, we design our patient-hospital assignment system to more efficiently assign patients to hospitals based on patients' preferred hospitals, patient travel distance costs, and hospital load balance.

\subsection{Individual Optimization Solution Results}

\begin{figure}[!ht]
  \includegraphics[width=\textwidth]{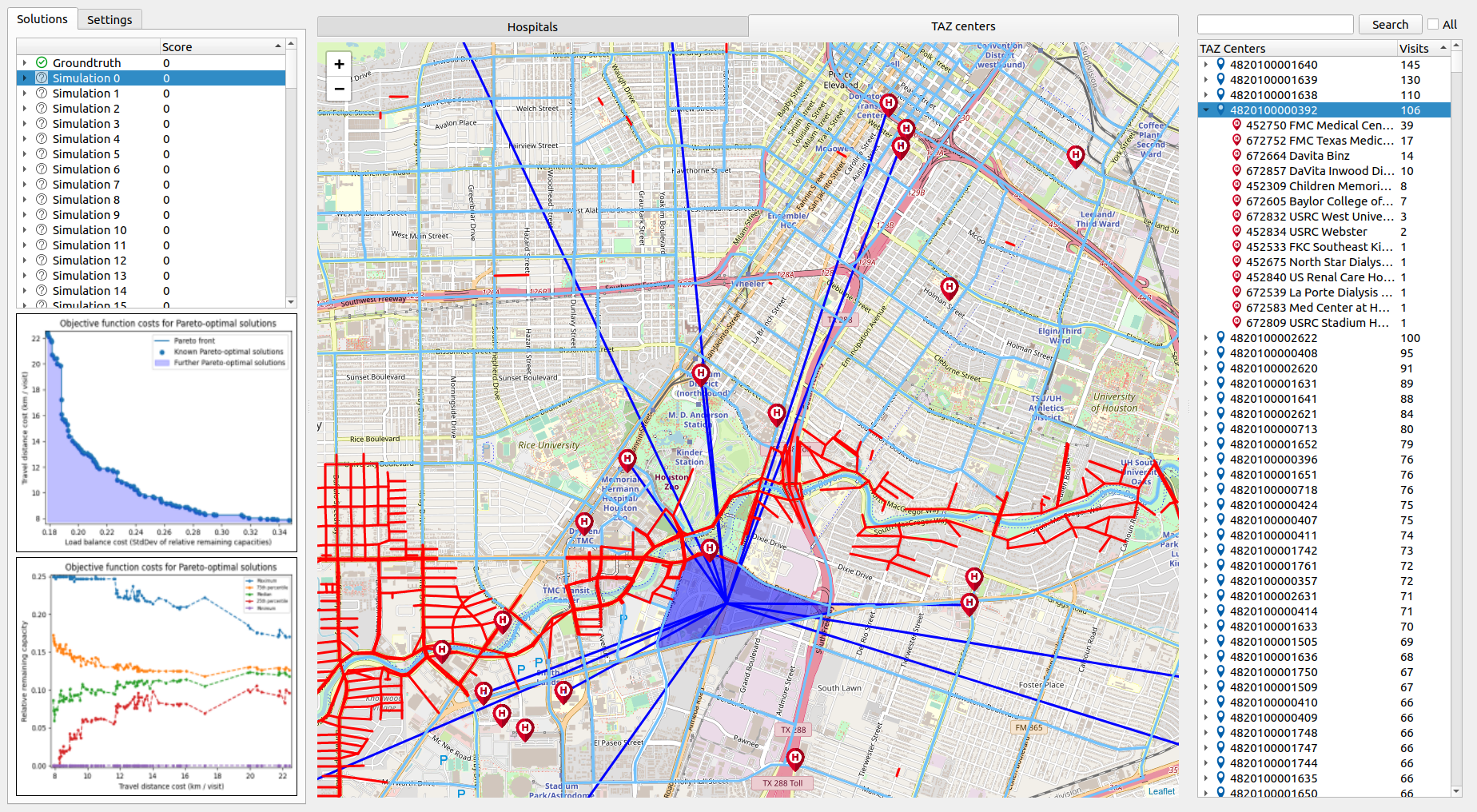}
  \caption{Interactive user interface for displaying scenario results. In the main map window, a TAZ (blue shaded region) sends its patients to hospitals (H-markers connected with a blue line) according to inundation conditions (inundated roads in red).}
  \label{fig:gui}
\end{figure}

The algorithm optimizes for minimizing travel distance cost and balancing load across hospitals. A solution for a scenario is a discrete assignment of patients from TAZs to hospitals. This patient-hospital assignment is not only evaluated by its objective function scores, but may also be judged through supplementary metrics determined by the solution. We also create an interactive user interface to graphically display the results of each scenario (\autoref{fig:gui}); we generate hospital statuses and TAZ statuses as described in \ref{subsection:postprocess} and display them in the interface.

\begin{figure}[!ht]
    \centering
    \begin{subfigure}{0.49\textwidth}
        \centering
        \includegraphics[width=0.99\linewidth]{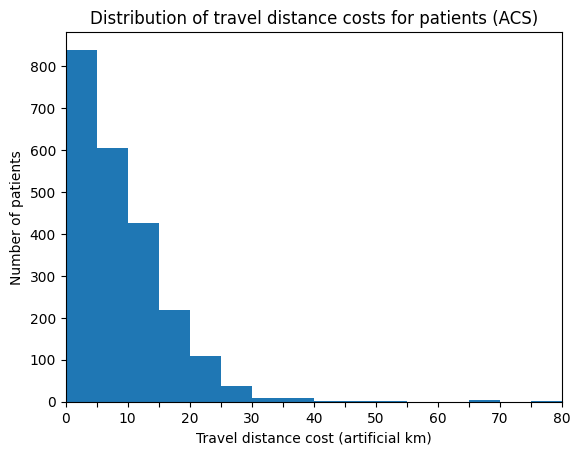}
        \caption{Travel distance costs for each patient}
        \label{fig:00d}
    \end{subfigure}
    \begin{subfigure}{0.49\textwidth}
        \centering
        \includegraphics[width=0.99\linewidth]{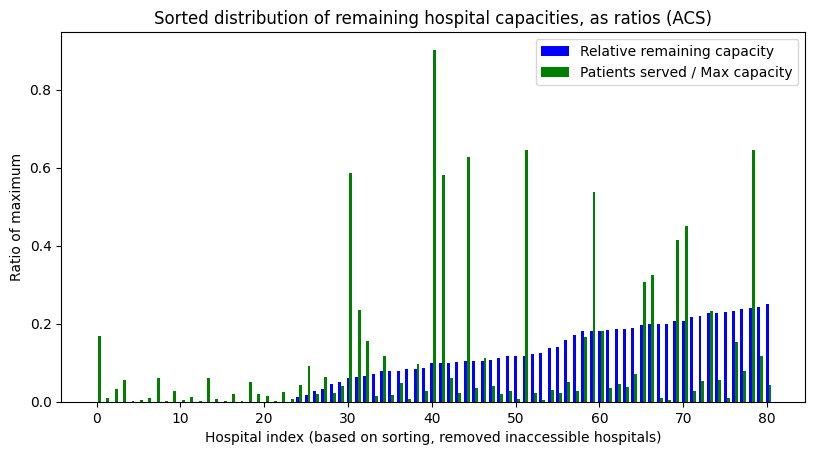}
        \caption{Normalized count of patients served (indicating size of hospital; in green) and relative unused capacity for each hospital (1 - occupancy ratio; in blue)}
        \label{fig:00lb}
    \end{subfigure}
    \caption{Scenario 00 - Solution with minimum distance cost}
    \label{fig:00obj}
\end{figure}

The following two plots describe the resultant distributions related to the objective functions. In \autoref{fig:00d}, we see the distribution of travel distance costs for re-assigned patients according to the solution with minimum distance cost for scenario 00. In \autoref{fig:00lb}, we see the normalized number of patients served by each hospital, with hospitals sorted in order of increasing relative unused capacity. We see that this solution (which has minimum distance cost, with an average of $\approx 8.94$km per visit) has the vast majority of patients traveling under (20) kilometers, with a small proportion of patients traveling up to $\approx 80$ kilometers. The hospital load balance is not ideal as a result of prioritizing travel distance, as hospitals are unevenly burdened. On the other hand, it is seen that there are no large hospitals which are being pushed to, let alone beyond, capacity, since our optimization framework prevents exceeding capacity.

\begin{figure}[!ht]
    \centering
    \begin{subfigure}{0.49\textwidth}
        \centering
        \includegraphics[width=0.99\linewidth]{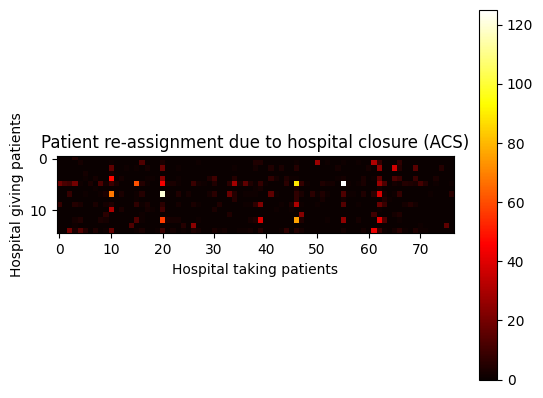}
        \caption{Condensed hospital-to-hospital patient re-assignment matrix (rows and columns of zeros removed)}
        \label{fig:00hhm2}
    \end{subfigure}
    \begin{subfigure}{0.49\textwidth}
        \centering
        \includegraphics[width=0.99\linewidth]{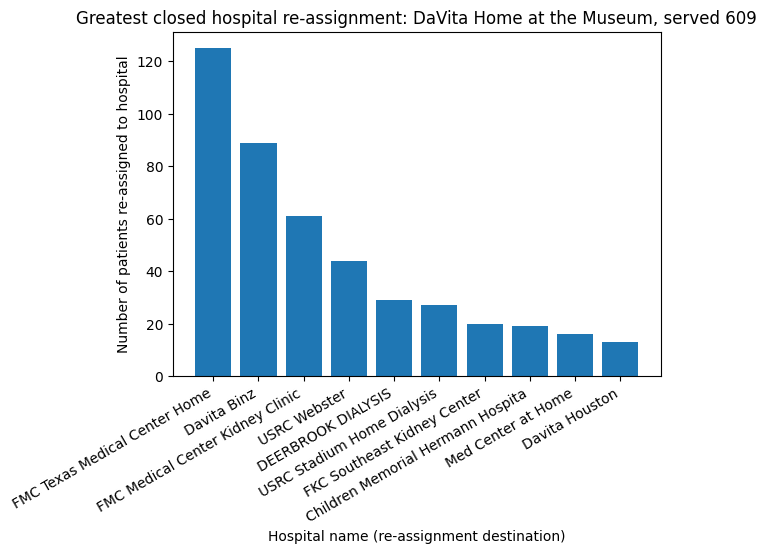}
        \caption{Greatest closed hospital re-assignment, Scenario 00, minimum distance solution}
        \label{fig:00gchm}
    \end{subfigure}

    \caption{Scenario 00 - Hospital-to-hospital patient re-assignment, minimum distance solution}
    \label{fig:00hhm}
\end{figure}

In \autoref{fig:00hhm2}, we see a visual, one-to-one representation of the hospital-to-hospital patient re-assignment matrix for the same solution and scenario. This matrix describes the number of patients which were reassigned to each hospital from the hospital each patient normally visits but is closed within this scenario. An active row describes a hospital giving its patients due to closure, and an active column describes a hospital receiving patients from closed hospitals. In this case, 15 of 95 hospitals are closed, and 80 of 95 hospitals are open; typically all or nearly all open hospitals will receive at least one new patient.

Looking at a subset of the data from another perspective, we can view the top (10) hospitals in terms of number of patients that were re-assigned from the closed hospital which was serving the greatest number of patients. In \autoref{fig:00gchm}, we see that the hospital ``FMC Texas Medical Center Home'' received approximately one fifth (120+) of the patients from the greatest closed hospital ``DaVita Home at the Museum'', which served (609) patients originally.

These individual solution results are used in aggregate to inform scenario-wide results (\ref{subsection:scenario-wide}) and aggregate results (\ref{subsection:all_scenario}), which lead to key insights (\ref{subsection:key_points}).

\subsection{Scenario-wide Pareto Set \& Scenario-wide Aggregate Data} \label{subsection:scenario-wide}
\begin{figure}[!ht]
    \centering
    \begin{subfigure}{0.49\textwidth}
        \centering
        \includegraphics[width=0.99\linewidth]{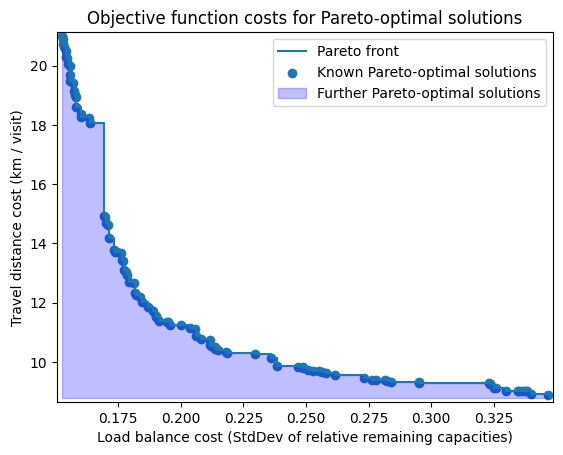}
        \caption{Pareto front}
        \label{fig:00p}
    \end{subfigure}
    \begin{subfigure}{0.49\textwidth}
        \centering
        \includegraphics[width=0.99\linewidth]{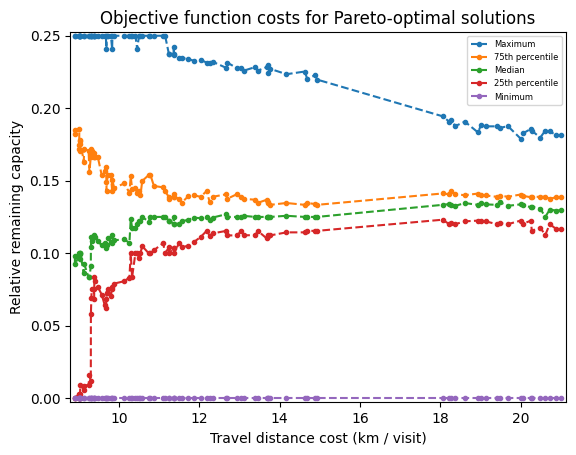}
        \caption{Hospital relative unused capacity spread vs. distance cost}
        \label{fig:00p2}
    \end{subfigure}
    \caption{Scenario 00 - Pareto set visualizations}
    \label{fig:00pa}
\end{figure}

Each globally Pareto-optimal solution set (one from each run of the algorithm on each scenario) is merged into a single universally Pareto-optimal solution set. This Pareto set may be visualized in the following two ways; \autoref{fig:00p} shows the Pareto front, and \autoref{fig:00p2} shows the decrease in spread of relative unused capacity at various percentiles for hospitals as greater and greater travel distance costs are allowed.

This Pareto front plot (\autoref{fig:00p}) shows (93) distinct solutions. The horizontal axis measures the hospital load balance cost. The vertical axis measures the patient travel distance cost. The algorithm is fully capable producing a wide range of solutions along the Pareto front, which shifts some preference for one objective function to the other. In this instance, the maximum load balance cost is $\approx 227\%$ of the minimum, and the maximum travel distance cost is $\approx 236\%$ of the minimum. 

\subsubsection{Effects of Heuristic Function Choice on Objective Function Distribution}
We showed in \ref{subsection:scenario-wide} that our process gives many Pareto-optimal solutions with a wide range of solution choices to cater towards a specific preference weighting of objective functions. The two objective functions are naturally opposing in practice, and figure \autoref{fig:00p2} visually represents a clear trade-off between travel distance cost and balancing hospital load. A major contributing factor as to how this is able to be achieved is the use of multiple runs of the algorithm on each scenario. Specifically, each run differs in its combination of $\gamma_0$ and $\gamma_1$ (\ref{subsection:h}), which determine the shape of the heuristic function curves, and therefore significantly alter the character of the resultant distributions of the metric each heuristic relates to. 

\begin{figure}[!ht]
    \centering
    \begin{subfigure}{0.49\textwidth}
        \centering
        \includegraphics[width=0.99\linewidth]{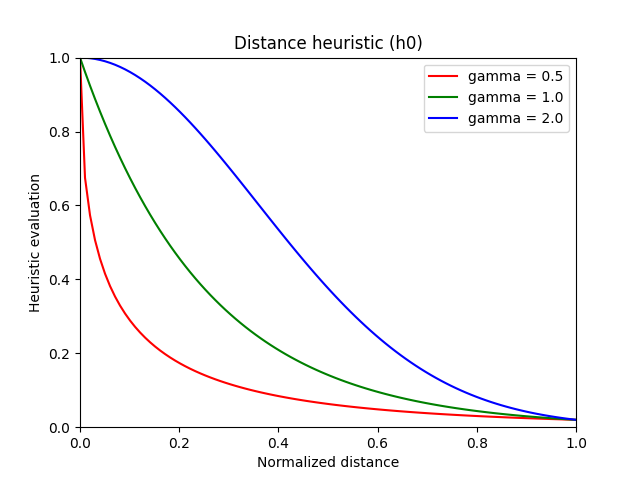}
        \caption{Distance heuristic}
        \label{fig:hdg1}
    \end{subfigure}
    \begin{subfigure}{0.49\textwidth}
        \centering
        \includegraphics[width=0.99\linewidth]{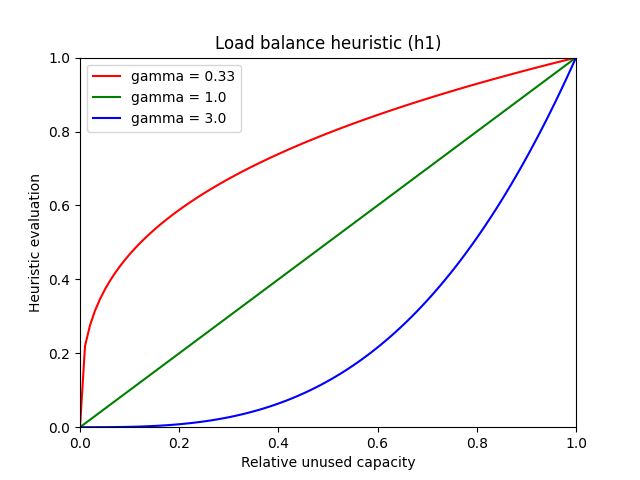}
        \caption{Load balance heuristic}
        \label{fig:hdg2}
    \end{subfigure}
    \caption{Heuristic functions and choices for $\gamma$}
    \label{fig:hdg0}
\end{figure}

First, we provide some level of intuition about how each $\gamma$ affects the heuristic function itself. Shown in \autoref{fig:hdg1} is the heuristic relating to distance cost $h_0$, for each of the different choices of $\gamma_0 \in \{\frac{1}{2}, 1, 2\}$. Shown in \autoref{fig:hdg2} is the heuristic relating to distance cost $h_1$, for each of the different choice of $\gamma_1 \in \{\frac{1}{3}, 1, 3\}$. 

Note how for $h_0$, a lower value of $\gamma_0$ heavily discourages anything but distances very close to zero; and, a high value of $\gamma_0$ treats lesser distances similarly to each other and greater distance similarly to each other. On the other hand, the effect of $\gamma_1$ is more straightforward: a lesser $\gamma_1$ devalues the importance of the relative unused capacity, whereas a greater $\gamma_1$ amplifies its importance. 

\begin{figure}[!ht]
    \centering
    \begin{subfigure}{0.32\textwidth}
        \centering
        \includegraphics[width=0.99\linewidth]{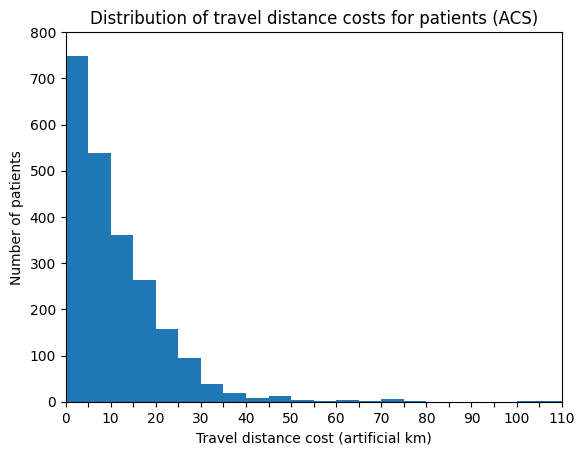}
        \caption{$(\gamma_0, \gamma_1) = (\frac{1}{2}, 3)$}
        \label{fig:heuristic_distance_effect1}
    \end{subfigure}
    \begin{subfigure}{0.32\textwidth}
        \centering
        \includegraphics[width=0.99\linewidth]{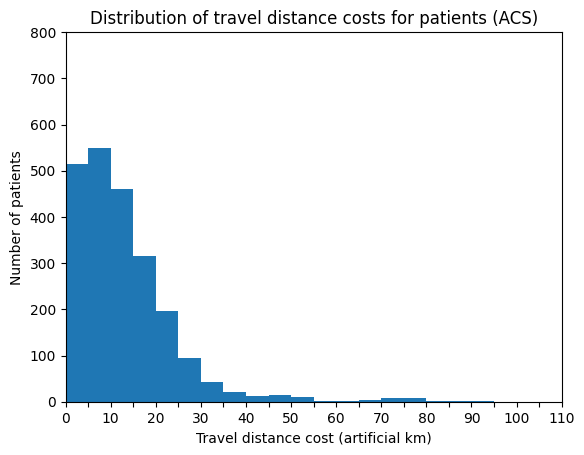}
        \caption{$(\gamma_0, \gamma_1) = (1, 3)$}
        \label{fig:heuristic_distance_effect2}
    \end{subfigure}
    \begin{subfigure}{0.32\textwidth}
        \centering
        \includegraphics[width=0.99\linewidth]{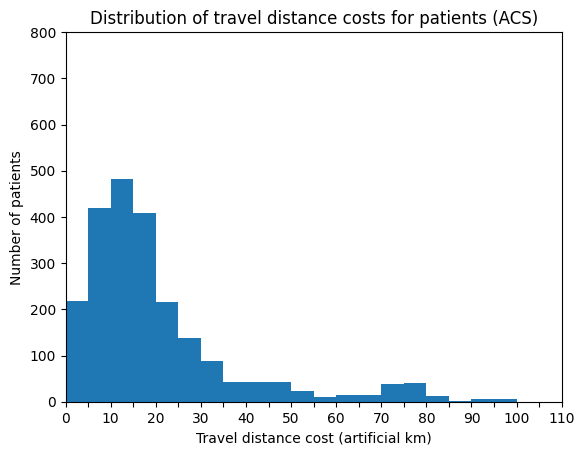}
        \caption{$(\gamma_0, \gamma_1) = (2, 3)$}
        \label{fig:heuristic_distance_effect3}
    \end{subfigure}
    \caption{Patient travel distance cost distribution for varied $\gamma_0$, Scenario 00}
    \label{fig:heuristic_distance_effect}
\end{figure}

In \autoref{fig:heuristic_distance_effect}, we see the travel distance cost distribution for the first solution from each of three runs varying $\gamma_0$ for scenario 00: $(\gamma_0, \gamma_1) \in \{(\frac{1}{2}, 3), (1, 3), (2, 3)\}$. As expected, the least value of $\gamma_0$ of these three corresponds to the least distance objective cost of these three, and likewise for the greatest value of $\gamma_0$. However, further notice that $\gamma_0 = 2$ gives a differently shaped curve compared to $\gamma_0 \in \{\frac{1}{2}, 1\}$, whose heuristic curves themselves look similar.

\begin{figure}[!ht]
    \centering
    \begin{subfigure}{0.32\textwidth}
        \centering
        \includegraphics[width=0.99\linewidth]{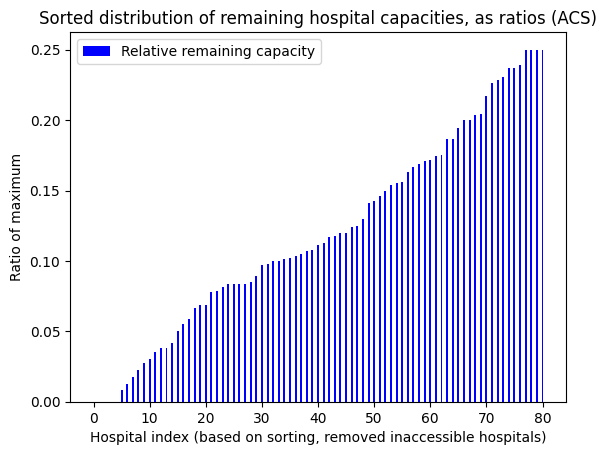}
        \caption{$(\gamma_0, \gamma_1) = (1, \frac{1}{3})$}
        \label{fig:heuristic_load_balance_effect1}
    \end{subfigure}
    \begin{subfigure}{0.32\textwidth}
        \centering
        \includegraphics[width=0.99\linewidth]{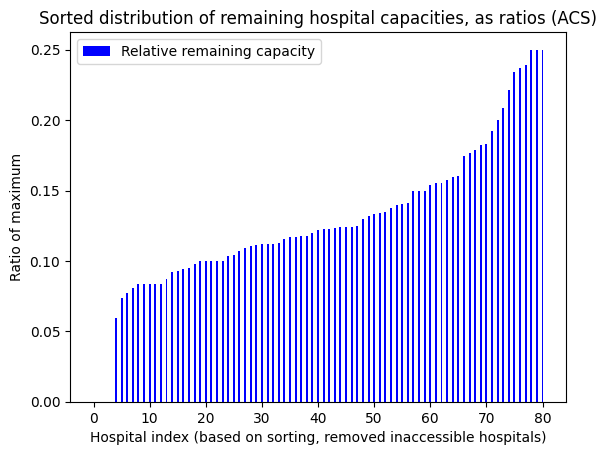}
        \caption{$(\gamma_0, \gamma_1) = (1, 1)$}
        \label{fig:heuristic_load_balance_effect2}
    \end{subfigure}
    \begin{subfigure}{0.32\textwidth}
        \centering
        \includegraphics[width=0.99\linewidth]{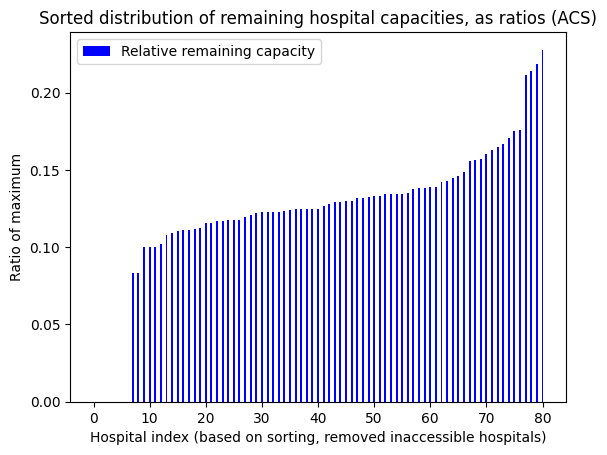}
        \caption{$(\gamma_0, \gamma_1) = (1, 3)$}
        \label{fig:heuristic_load_balance_effect3}
    \end{subfigure}
    \caption{Sorted relative unused capacity distribution for varied $\gamma_1$, Scenario 00}
    \label{fig:heuristic_load_balance_effect}
\end{figure}

In \autoref{fig:heuristic_load_balance_effect}, we see the natural effect of $\gamma_1$ in $(\gamma_0, \gamma_1) \in \{(1, \frac{1}{3}), (1, 1), (1, 3)\}$ on the load balance heuristic. This effect is more straightforward than that of $\gamma_0$; it is clearly seen that raising $\gamma_1$ has the direct consequence of reducing spread in relative unused capacity. 

Overall, this shows that the choice of heuristic function acts as a predictable, interpretable method of control for adjusting characteristics of the solution.

\subsection{All-scenario Aggregate Results}
\label{subsection:all_scenario}

From the all-scenario aggregate results, we may find the following statistics to generate insights (\ref{subsection:key_points}) about hurricane management and planning. In this aggregation process, each scenario and solution is weighted equally.

\begin{figure}[!ht]
    \centering
    \begin{subfigure}{0.49\textwidth}
        \centering
        \includegraphics[width=0.99\linewidth]{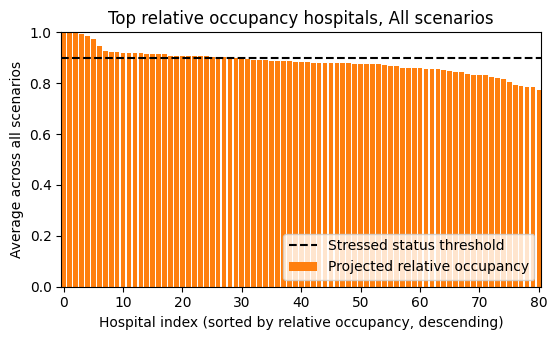}
        \caption{Hospitals sorted by relative occupancy}
        \label{fig:all_scenario_top_relative_occupancy}
    \end{subfigure}
    \begin{subfigure}{0.49\textwidth}
        \centering
        \includegraphics[width=0.99\linewidth]{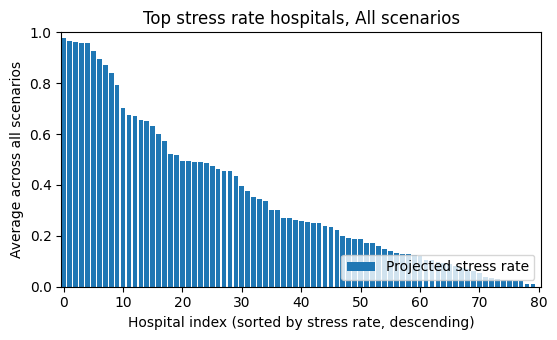}
        \caption{Hospitals sorted by stress rate}
        \label{fig:all_scenario_stress_rate}
    \end{subfigure}
    \caption{All-scenario aggregate results. Note that hospitals which are always closed are excluded in this plot.}
\end{figure}

In \autoref{fig:all_scenario_top_relative_occupancy}, we see the expected relative occupancy of each hospital sorted in descending order. It is seen that some hospitals are, in expectation, being stressed by the patient load (over $90\%$ occupancy). The hospital with the greatest average relative occupancy (``hospital 0'', the left-most bar in this plot) is ``Fresenius Kidney Care of Crosby''; the hospital with the least average relative occupancy (``hospital 80'', the right-most bar in this plot) is ``FMC Baytown Dialysis Facility''.

In \autoref{fig:all_scenario_stress_rate}, we see the projected rates at which each hospital is ``stressed'' (meaning it is over $90\%$ occupancy), with rates sorted in descending order. The hospital with the greatest stress rate (``hospital 0'', the left-most bar in this plot) is again ``Fresenius Kidney Care of Crosby''. The hospital with the least stress rate (``hospital 80'', the right-most bar in this plot) is ``Davita Baytown Dialysis'', the second largest hospital.

\begin{figure}[!ht]
    \centering
    \begin{subfigure}{0.49\textwidth}
        \centering
        \includegraphics[width=0.99\linewidth]{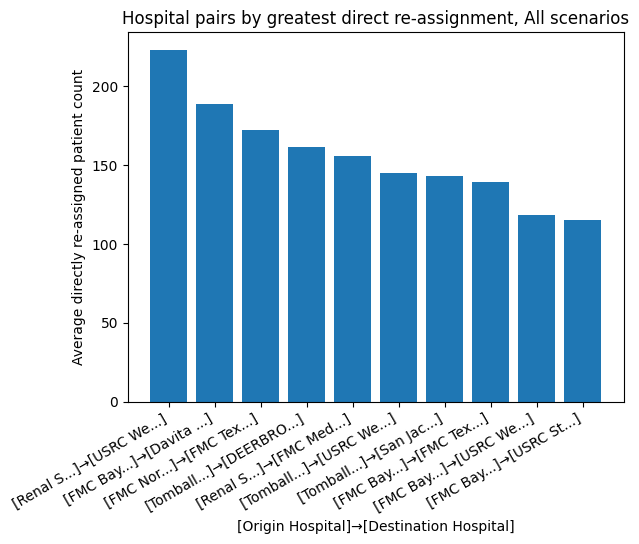}
        \caption{Hospital pairs by greatest direct re-assignment, All scenarios}
        \label{fig:all_scenario_top_direct_reassignment}
    \end{subfigure}
    \begin{subfigure}{0.49\textwidth}
        \centering
        \includegraphics[width=0.99\linewidth]{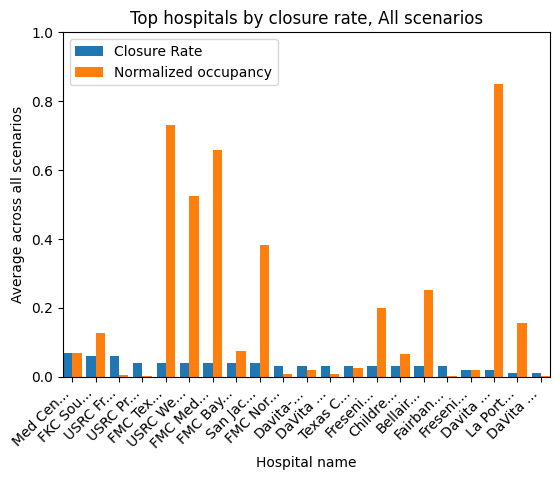}
        \caption{Most important hospitals at risk of closure, All scenarios}
        \label{fig:all_scenario_closed_hospital_importance}
    \end{subfigure}
    \caption{All scenarios - Hospital pairs by direct re-assignment and hospitals by closure rate.}
    \label{fig:all_scenarios_closure_results}
\end{figure}

In \autoref{fig:all_scenario_top_direct_reassignment}, we see the pairs of hospitals $(X, Y)$ with greatest direct re-assignment, such that when hospital $X$ closes, a significant number of its patients in expectation are re-assigned to hospital $Y$. 

In \autoref{fig:all_scenario_closed_hospital_importance}, we see the relative importance of the (21) hospitals which were at risk of closure (this excludes hospitals which were closed in all scenarios).

\begin{figure}[!ht]
    \centering
    \includegraphics[width=0.99\linewidth]{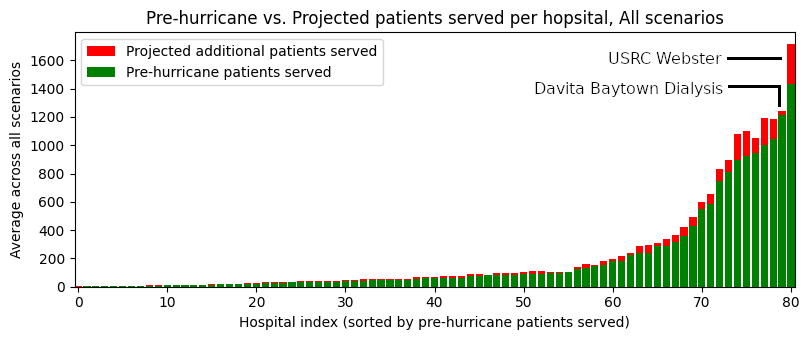}
    \caption{All scenarios - Count of patients served per hospital, pre-hurricane vs. projected}
    \label{fig:all_scenario_patients_served}
\end{figure}

In \autoref{fig:all_scenario_patients_served}, we see the number of patients each hospital served pre-hurricane compared to the projected during-hurricane estimate. This plot suggests that it is mainly the largest hospitals which are consistently taking in the greatest number of new patients during the hurricane, particularly the largest hospital, USRC Webster. It is seen that USRC Webster is projected to serve a significant number of additional patients, whereas Davita Baytown Dialysis (the second largest hospital) is not; this may suggest that USRC Webster is more advantageously located compared to Davita Baytown Dialysis. Note that in this plot, the (14) hospitals which are closed in every scenario are excluded. 

\begin{figure}[!ht]
    \centering
    \begin{subfigure}{0.49\textwidth}
        \centering
        \includegraphics[width=0.99\linewidth]{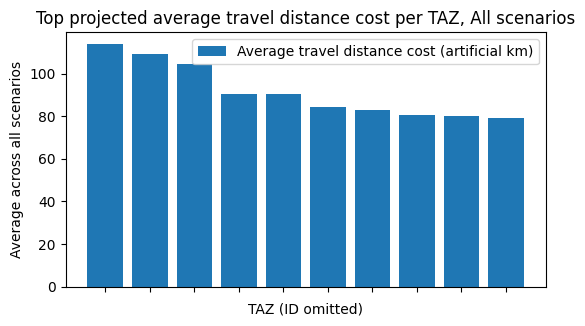}
        \caption{All scenarios, top TAZs by projected average travel distance}
        \label{fig:all_scenarios_top_taz_risk}
    \end{subfigure}
    \begin{subfigure}{0.49\textwidth}
        \centering
        \includegraphics[width=0.99\linewidth]{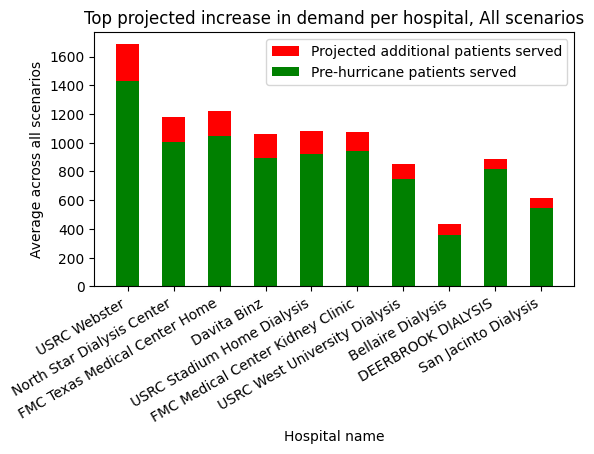}
        \caption{All scenarios, top hospitals by projected demand increase}
        \label{fig:all_scenarios_demand_increase}
    \end{subfigure}
    \caption{All scenarios - Top Pre-hurricane results vs. Projected results}
    \label{fig:all_scenarios_top_n_plots}
\end{figure}

Lastly, we examine the top (10) TAZs by expected travel distance cost in \autoref{fig:all_scenarios_top_taz_risk}. We examine the top (10) hospitals by greatest increase in demand from pre-hurricane visit count to average simulated visit count in \autoref{fig:all_scenarios_demand_increase}.

\subsubsection{Key Points of Recommendation}
\label{subsection:key_points}
The results described above demonstrate that our process is able to model and optimize the patient-hospital re-assignment problem. Our key points of recommendation are as follows:

\begin{itemize}
    \item We discern which TAZs are most at risk of needing to send its patients significantly farther in case of hurricane (\autoref{fig:all_scenarios_top_taz_risk}); these patients may be notified accordingly and prepare for the hurricane, perhaps with an in-home dialysis procedure or by scheduling a longer drive.
    \item We identify which hospitals, which if closed due to a hurricane, sees its patients reassigned to another hospital in the greatest quantity (\autoref{fig:all_scenario_top_direct_reassignment}); these hospital pairs may elect to share patient information, etc. in order to prepare for the transfer of patients and make their transition smoother.
    \item We identify certain key hospitals which are most critical to ensure facility reliability based on (\autoref{fig:all_scenario_patients_served}, \autoref{fig:all_scenarios_demand_increase}, \autoref{fig:all_scenario_closed_hospital_importance}).
    \item We identify certain key hospitals which would most benefit from an increase in capacity based on (\autoref{fig:all_scenario_top_relative_occupancy}, \autoref{fig:all_scenario_stress_rate}).
    \item Lastly, a city planning official or hospital system manager may choose to construct new hospitals near TAZs with a high projected travel distance cost (\autoref{fig:all_scenarios_top_taz_risk}) or near existing hospitals as an alternative way to increase the capacity of an often-stressed area (\autoref{fig:all_scenario_top_relative_occupancy}, \autoref{fig:all_scenario_stress_rate}). 
\end{itemize}

\section{Discussion}

The use of data-driven methods for disaster risk management can significantly reduce the impacts of hazards on communities (and vulnerable populations in particular). This study is a first of its kind to leverage location-based datasets along with datasets related to critical care facilities, road networks, and hazard exposures in creating and testing an optimization model to efficiently re-distribute dialysis-dependent patients when access to a number of facilities is disrupted during extreme weather events. The current approach to preparing and responding to disruptions in access to dialysis facilities is rather reactive and could lead to sub-optimal allocation of patients to other facilities. Through the use of the proposed data-driven optimization model, public health officials and emergency managers could proactively evaluate different scenarios of road inundations and facility outages to identify: (1) areas most vulnerable to loss of access to dialysis centers; (2) optimal ways to allocate patients to dialysis centers for different scenarios individually and collectively; (3) capacity utilization of dialysis centers and facilities that play a critical role in the overall redundancy of the network of facilities within a city or region. These insights could inform plans to reduce the impacts of disrupted access by increasing the capacity of most important facilities during extreme weather events, building new facilities in areas with the greatest vulnerability, and reducing vulnerability of the existing facilities through backup power and water supply. These proactive measures obtained based on data-driven methods could prevent catastrophic kidney failure disasters in the face of growing extreme weather events. Also, the method and data presented in this study could be used for optimizing access to other critical care facilities beyond dialysis centers. 

Incorporating data-driven methods into disaster risk management has the potential to significantly mitigate the adverse impacts of hazards on communities, especially the most vulnerable. This study introduces a breakthrough approach by utilizing location-based datasets, critical care facility information, road networks, and hazard exposure data to develop and validate an optimization model. This model efficiently redistributes dialysis patients during disruptive weather events, addressing a challenge that requires proactive solutions. Traditionally, responses to access disruptions have been reactive and suboptimal, potentially compromising patient care.

In contrast, our data-driven optimization model enables public health officials and emergency managers to anticipate and strategize effectively. They can evaluate different scenarios of road inundations and facility closures to identify at-risk areas, optimize patient distribution, and enhance the redundancy of care networks. These insights can inform plans to reduce the impact of disrupted access by increasing the capacity of facilities in critical areas during extreme weather events and establishing new facilities in vulnerable zones. By integrating data-driven foresight, we can prevent catastrophic incidents of kidney failure during extreme weather events.

Moreover, the method and dataset presented here have broader applications, extending to the optimization of various critical care facilities beyond dialysis centers. The combination of data-driven innovation, strategic planning, and healthcare resilience marks a significant shift in disaster preparedness and response. With this comprehensive approach, we can protect communities, save lives, and strengthen the integrity of critical healthcare systems.

\section{Conclusion}

Our methodology as presented in this work demonstrates the concept of applying a multi-objective optimization algorithm to a real-world data-based model of flooded roads and closed dialysis centers due to a hurricane, and re-assigning dialysis patients from closed dialysis centers to operational dialysis centers in an organized, centralized manner. From our hurricane flooding simulation scenario results, we provide recommendations for Houston, Texas city planning officials and hospital system coordinators; we show TAZs in which patients are most at risk of having a high travel burden to receive care, which hospitals are most critical to ensure facility reliability, which hospitals give the most benefit from an increase in capacity, and where new hospitals may be constructed to provide the greatest system resilience. 

\section{Glossary \& Abbreviations}
\begin{itemize}
    \item TAZ: Traffic Analysis Zone
    \item ACO: Ant Colony Optimization
    \item ACS: Ant Colony System
\end{itemize}

\section{Acknowledgements \& Funding sources}

This work was supported by Defense Advanced Research Projects Agency (DARPA) under contract number W31P4Q-18-C-0055. The views, opinions and/or findings expressed are those of the author and should not be interpreted as representing the official views or policies of the Department of Defense or the U.S. Government.

\section{Distribution Statement}
“A” - Approved for Public Release, Distribution Unlimited.

\section{Author contributions: CRediT}

\textbf{Emmanuel Tung:} Conceptualization; Data Curation; Formal Analysis; Investigation; Methodology; Software; Validation; Visualization; Writing – Original Draft Preparation; Writing – Review \& Editing. 

\textbf{Ali Mostafavi:} Conceptualization; Data Curation; Methodology; Resources; Supervision; Writing – Original Draft Preparation; Writing – Review \& Editing. 

\textbf{Maoxu Li:} Data Curation; Software; Visualization. 

\textbf{Sophie Li:} Data Curation; Software. 

\textbf{Zeeshan Rasheed:} Conceptualization; Funding Acquisition; Project Administration; Resources; Supervision; Visualization; Writing – Review \& Editing.

\textbf{Khurram Shafique:} Conceptualization; Funding Acquisition; Project Administration; Resources; Supervision; Writing – Review \& Editing.

\section{Data statement}

The authors do not have permission to share data. 

\section{Declaration of interests}

The authors declare that they have no known competing financial interests or personal relationships that could have appeared to influence the work reported in this paper.

\appendix
\section{Supplementary Materials}
\label{app1}

Values for algorithm parameters:

\begin{itemize}
    \item $n_i = 30$ (iteration count)
    \item $n_a = 30$ (ant count)
    \item $q_0 = 0.8$ (exploitation rate)
    \item $\alpha_l = 0.1$ (local pheromone decay rate)
    \item $\alpha_g = 0.1$ (global pheromone decay rate)
    \item $\beta = 2.0$ (heuristic importance factor)
\end{itemize}

\bibliographystyle{elsarticle-harv} 
\bibliography{main}

\end{document}